\documentclass[
]{ceurart}

\usepackage{caption}
\usepackage{balance}
\usepackage{caption}

\usepackage{makecell}

\usepackage{graphicx} 
\graphicspath{{pictures/}}
\DeclareGraphicsExtensions{.pdf,.png,.jpg}

\begin{document}

\copyrightyear{2021}
\copyrightclause{Copyright for this paper by its authors.
  Use permitted under Creative Commons License Attribution 4.0
  International (CC BY 4.0).}

\conference{GraphiCon 2021: 31th International Conference on Computer Graphics and Vision,
  September 27--30, 2021, Nizhny Novgorod, Russia}

\title{Shot boundary detection method based on a new extensive dataset and mixed features}

\author[1]{Alexander Gushchin}[%
email=alexander.gushchin@graphics.cs.msu.ru
]

\author[1]{Anastasia Antsiferova}[%
email=aantsiferova.graphics.cs.msu.ru
]
\address[1]{Lomonosov Moscow State University,
  119991, Russia, Moscow, Leninskiye Gory, 1}

\author[1]{Dmitriy Vatolin}[%
email=dmitriy@graphics.cs.msu.ru
]
\begin{abstract}
Shot boundary detection in video is one of the key stages of video data processing. A new method for shot boundary detection based on several video features, such as color histograms and object boundaries, has been proposed. The developed algorithm was tested on the open BBC Planet Earth \cite{bbc} and RAI \cite{rai} datasets, and the MSU CC datasets, based on videos used in the video codec comparison conducted at MSU, as well as videos from the IBM set, were also plotted. The total dataset for algorithm development and testing exceeded the known TRECVID datasets. Based on the test results, the proposed algorithm for scene change detection outperformed its counterparts with a final F-score of 0.9794.\end{abstract}


\maketitle

\section{Introduction}

\begin{figure*}[htb]
\centering
\includegraphics[width=\linewidth]{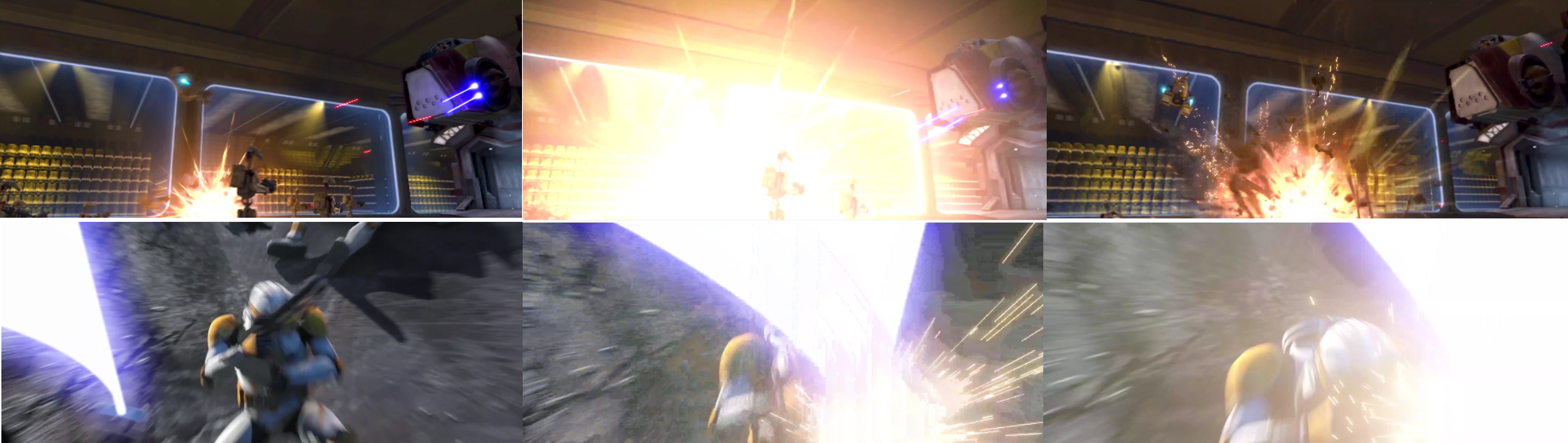}
\caption{Examples of flashes which cause errors in shot boundary detection.}
\label{fig:flashes_example}
\end{figure*}

One of the basic steps in video processing is video scene splitting. For example, scene cutting is a necessary step in video annotation and indexing \cite{first}, keyframe searching \cite{second}, and automatic video format changing \cite{third}.
Existing algorithms have achieved high accuracy in detecting transitions between scenes in general cases, but still make mistakes in detecting complex transitions (Fig.~\ref{fig:flashes_example}).

Also, existing algorithms have been developed based on open data sets that may contain errors. For example, when analyzing one of the most popular BBC \cite{bbc} datasets, several frame inaccuracies were found in the markup of transitions: for example, in the From Pole to Pole video, the first scene ends at 632 frames, but the second begins at 650 frames. Since the algorithm must specify the frame number with the scene change, such errors have been corrected to work more accurately.
Thus, the challenges of creating a method for quickly and accurately partitioning video into scenes, as well as creating a volumetric data set with accurate partitioning, are relevant.

\begin{figure*}[htb]
\center{\includegraphics[width=.75\linewidth]{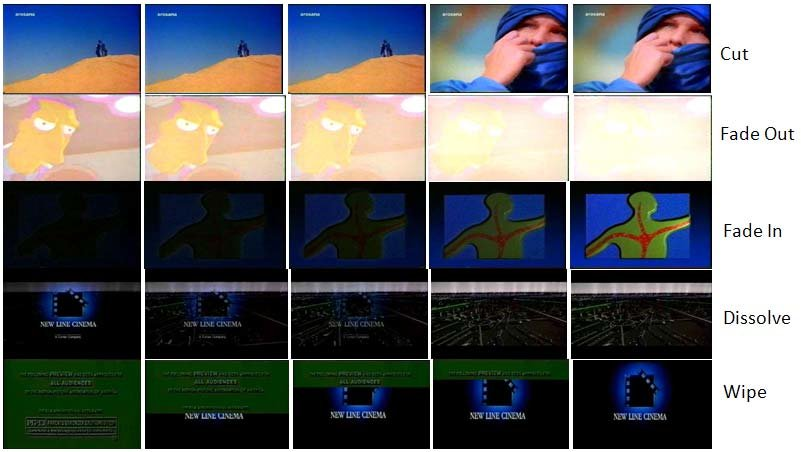}}
\caption{Examples of scene changes.}
\label{fig:image33}
\end{figure*}

Since different definitions for scene transitions are found in the literature (for example, the definition of a scene varies -- it can be defined as a gluing of two perspectives or a semantic part of a movie), the following are the basic definitions that will be used in this work. In this paper, we have relied on the definitions given in the formal statement of the problem formulated by the authors in \cite{formal}. 
The basic element of video is a frame, frames are combined into scenes (shots), and scenes are combined into semantic scenes. A scene---is a continuous stretch of video, shot with a single camera, without stitches or interruptions. A semantic scene is a sequence of scenes with the same semantics.
The task of the shot boundary detector is to indicate all the frames in which a scene change has occurred. In most cases, the content within a scene changes gradually, and at the boundaries there is montage gluing, so this task is trivial for humans.
Scene changes themselves are divided into two types -- abrupt and gradual. Abrupt changes in scenes - the momentary transition from a frame of one scene to a frame of the next. This can be dissolve (the gradual appearance of a new scene on top of the previous one), fade (a gradual transition to a black frame and back) or wipe. Examples of such transitions are shown in Fig.~\ref{fig:image33}.

Most shot boundary detection algorithms work in 2 steps:
\begin{itemize}
\item Calculating the value of the frame difference metric or metrics
\item Setting the threshold for frame classification. Also at this stage, machine learning is often used for automatic classification.
\item An additional step can be filtering frames for false positive detections.
\end{itemize}

The purpose of this work was to create a new method for shot boundary detection and compare it with existing methods on a new large volume of data. The paper is further structured as follows:
\begin{itemize}
\item In section 2, an overview of algorithms from the field is given, as well as datasets to compare them
\item Section 3 gives a detailed overview of the proposed approach to solve the problem at hand
\item Section 4 contains the results of testing the open algorithms and comparing them with the proposed method
\item Section 5 contains conclusions and further plans.
\end{itemize}

\section{Related work}

In one of the most detailed works devoted to the analysis of \cite{methods} shot boundary detection algorithms, the authors considered their drawbacks as ways to improve them. The main drawbacks include the slow speed of operation, as well as errors in cases of flashes, fast camera movement, etc.

In most existing methods, the first step is the calculation of features for frames. One of the frequently used is the frame similarity metric for finding the degree of difference between frames. As the scene changes, the value of this metric will increase, while inside the scene it is close to zero. The most popular techniques are: calculation of color histograms \cite{methods}, \cite{aaa}, boundary gradients \cite{a} \cite{b} \cite{c}, geometric transformations \cite{f} \cite{fff}, motion vectors \cite{d} \cite{e}. One of the simplest methods of constructing this metric is a pixel-by-pixel comparison of frames \cite{methods}. Other difference metrics are also calculated between frames --- for example, \cite{bb} uses L*a*b* space and the formula for the distance between colors in it.

The construction of color histograms was used, for example, in \cite{methods}, \cite{aaa}. Histograms can be computed both for RGB and other color spaces (HSV, YCbCr, L*a*b*). With this approach, the algorithm is less sensitive to motion within the frame, but may produce many false positives for scenes with flashes and rapid light changes. The use of boundary gradients partially solves the problem of false positives when the camera or objects move within the frame, allowing you to use frame boundary matching without relying on lighting. Such a technique was used in \cite{a} and \cite{b}. In \cite{c}, the authors used object boundaries within frames to construct a histogram of directional gradients. The histograms for different levels of the resolution pyramid are concatenated. This approach allowed the authors to obtain the characteristics of object boundaries in the frame at different levels.

The motion vectors were used in the following works: \cite{dd} \cite{d} \cite{e}. By using them, the scene change detection algorithm can be adapted to the movement within the scene, the camera movement or the appearance of a large object in the frame. Vectors take longer to compute than approaches based on color or border histograms, but they can be used together with fast computable metrics and achieve high accuracy (for example, if we consider motion vectors before comparing frame boundaries). Also one of the popular techniques is geometric transformations of frames -- Contourlet, Fourier transform \cite{f}, Walsh Hadamard \cite{fff}. These methods are sensitive to frame motion and resolution, which can seriously increase the running time of the algorithm. The authors from [Contourlet] used an improved contourlet, which is not sensitive to the problems mentioned above. Rarer ways of constructing metrics include SIFT, SURF, entropy \cite{g} \cite{h}. They can give comparable accuracy, but require more computational resources.

Many of the methods described above can be applied not to the whole image, but to a part of it. In this approach frame is divided into blocks (overlapping or not) and metric is calculated for each block. Vector of such metrics can be concatenated, histograms (including cumulative ones) \cite{b}) or use statistics (e.g., expectation and variance). Partitioning into blocks allows the algorithm to be less susceptible to changes in certain parts of the scene (e.g., rapid movement of objects or flashes). Thus, using a combination of features based on different characteristics of video and individual frames allows to achieve higher detection accuracy, reduce the number of false positives, but increases the runtime of the algorithm.

After calculation of frame similarity metrics, each frame is classified into one of three categories: abrupt scene change (cut), gradual scene change (dissolve, vipe, fade in or fade out, no scene change. Since the algorithm needs to analyze all frames, which is a resource-intensive task, some authors use video preprocessing: they select, using additional fast algorithms, the segments where scene changes are supposedly present and further process only those segments. This approach assumes that no scene changes occur on frames that are not in these segments. There are three main approaches to classification: classification by threshold, adaptive threshold, and machine learning. Threshold (or a set of thresholds) is the simplest way to classify. The values of a metric or metrics are compared to a predetermined threshold and a decision is made as to whether a class belongs to a certain class. This approach is rarely used, as it is more advantageous to select thresholds for each individual video based on its features. The adaptive threshold does not have this disadvantage and can not only change depending on video \cite{b}, but also depending on metrics values in some neighborhood of the frame \cite{j}. Thus, the threshold is adjusted not for the whole video, but for a particular scene. Recently, due to the development of machine learning algorithms, they are increasingly used for classification: SVM\cite{k}, bagged tree classifier\cite{c}, k nearest neighbors, neural networks. The authors of \cite{trecvid} have analyzed the techniques used--- according to their research, SVM showed the best results.

Certain video artifacts significantly complicate the detection algorithms--for example, flashes and camera/object motion. Some of them can be eliminated at preprocessing stage --- for example, separate metrics for flashes \cite{l} are introduced. 

\subsection{Existing datasets and benchmarks}

In most papers on scene-shift detection methods, the authors compare the performance of algorithms on the dataset used in the TRECVID competition. This is one of the most famous and extensive comparisons of shot boundary detection methods, which has been conducted annually for 7 years. It tested 57 algorithms using different sets of marked videos. After the end of each competition, articles were released analyzing the participants' solutions and their results (e.g., \cite{trecvid}). The dataset included mostly documentaries and television shows. There are also a number of articles comparing shot boundary detection methods (e.g., \cite{methods}).

We requested access to the TRECVID dataset, but unfortunately, due to Covid-19, the authors were unable to provide it. (The vendor agreement requires sending the dataset on DVDs, and the authors can't do shipments at this time.) An alternative to this dataset is the use of public datasets. There are several datasets at the moment -- their comparison is given in the table. The BBC Planet Earth dataset consists of documentaries, RAI --- from 10 randomly selected segments of the broadcast of the television channel RAI scuola. These are mostly talk shows and documentaries. Clipshots includes 20 categories of videos (sports, animals, amateur ...) collected from youtube and weibo. The TRECVid dataset most commonly used for method testing is a collection of 7 datasets that have been selected from different video categories.
Unfortunately, the size of the available datasets is inferior to the TRECVID set. In this paper, a new dataset was created to train and test a new method for scene change detection (Table~\ref{tab:datastat}).

\begin{center}
\begin{table}
\caption{Sizes of datasets used to analyze metrics and create a new SBD method.}
  \label{tab:datastat}
\begin{tabular}{ccccc} 
 \toprule
 Dataset name & {Length (minutes)} & {Scene changes}\\ 
 \midrule
 Rai \cite{rai} & 93 & 985\\ 
 \hline
 \makecell{BBC Planet \\Earth \cite{bbc} } & 539 & 4844 \\ 
 \hline
 \makecell{MSU CC \\(manually \\marked up)} & 39 & 274 \\ 
 \hline
 \makecell{OS VSD \\(manually \\marked up)} & 954 & 6871\\ 
 \hline
 TRECVID 2007 & 360 & 2317 \\ [1ex] 
 \bottomrule
\end{tabular}
\end{table}
\end{center}

\section{Proposed method}

\subsection{Marking up a new dataset}
To create an algorithm for detecting scene shifts, a set of OS VSD \cite{osvsd} data was collected using Yandex.Toloka \cite{toloka}. 
The creation of a dataset is divided into several steps:
\begin{itemize}
\item  A few algorithms configured in a way to maximize the completeness of the results was running on all videos
\item  A list of potential scene changes was created by combining the results of all algorithms
\item  Each potential scene change was cut from the original video as a short video sequence of 40 frames long
\item  Yandex.toloka was used to show peoples all these sequences for markup
\item  \begin{itemize}
\item  For each video segment, observers indicated whether there was a scene change in it
\item  Each video segment was shown at least 5 different people, if the results were not unambiguous the number of observers increased until an agreement between observers was reached
\end{itemize}
\end{itemize}
This resulted in an additional 19 videos with a total duration of 965 minutes surpassing the existing TRECVID. The table above shows other comparative characteristics of this set.

\subsection{Base features}

Methods that have shown high accuracy in existing comparisons were used as base features for the new algorithm. To analyze them, a newly created OS VSD dataset was used, on which these methods were compared. 

In the first step, the proposed algorithm uses several metrics to describe frame differences. These metrics are built on a boundary gradient, a frame color histogram to describe frame differences. This approach allows to take into account several factors possible when changing scenes and to get more information about the frames being compared.
On the second stage, lgbm algorithm is applied to these metrics for classification. It was chosen as a result of experiments with different machine learning techniques.

First, let us describe the features that our algorithm relies on.
\begin{itemize}
\item Metrics proposed in \cite{dr}
 \begin{itemize} \item These metrics use the average value and standard deviation of the brightnesses of the pixels in the block of frames. \end{itemize}
\item Cumulative color histogram metric
 \begin{itemize} \item It is based on \cite{b}. First, the Sobel operator is applied to the frames to find the borders, the trapezoidal smoothing function is applied, and the cumulative histogram of frame blocks is calculated.  \end{itemize}
\item Metric proposed in the Max Remain repository\cite{max}
 \begin{itemize} \item It calculates the difference between color histograms of two consecutive frames and builds the difference between them. The output is a vector of length $3*n$, $n$--- number of columns in histogram. \end{itemize}
\item Histogram of borders of objects in the frame
 \begin{itemize} \item At the beginning we apply Sobel operator to find the borders, divide the frame into 100 non-overlapping blocks, build a histogram of borders and compare it with neighboring frames. A threshold is applied to cut off blocks which are different in neighboring frames. \end{itemize}
\item Metric proposed in the aysebilgegunduz\cite{ays} repository
 \begin{itemize} \item metric is the distance bhattacharyya between histograms of consecutive frames. \end{itemize}
\item metric of the PyScene algorithm\cite{pyscene}
\end{itemize}

The \cite{murashko} features were also tested, but were discarded during the experiments due to their low accuracy compared to the other metrics.

\subsection{Training}
\begin{figure}[htb]
\center{\includegraphics[width=\linewidth]{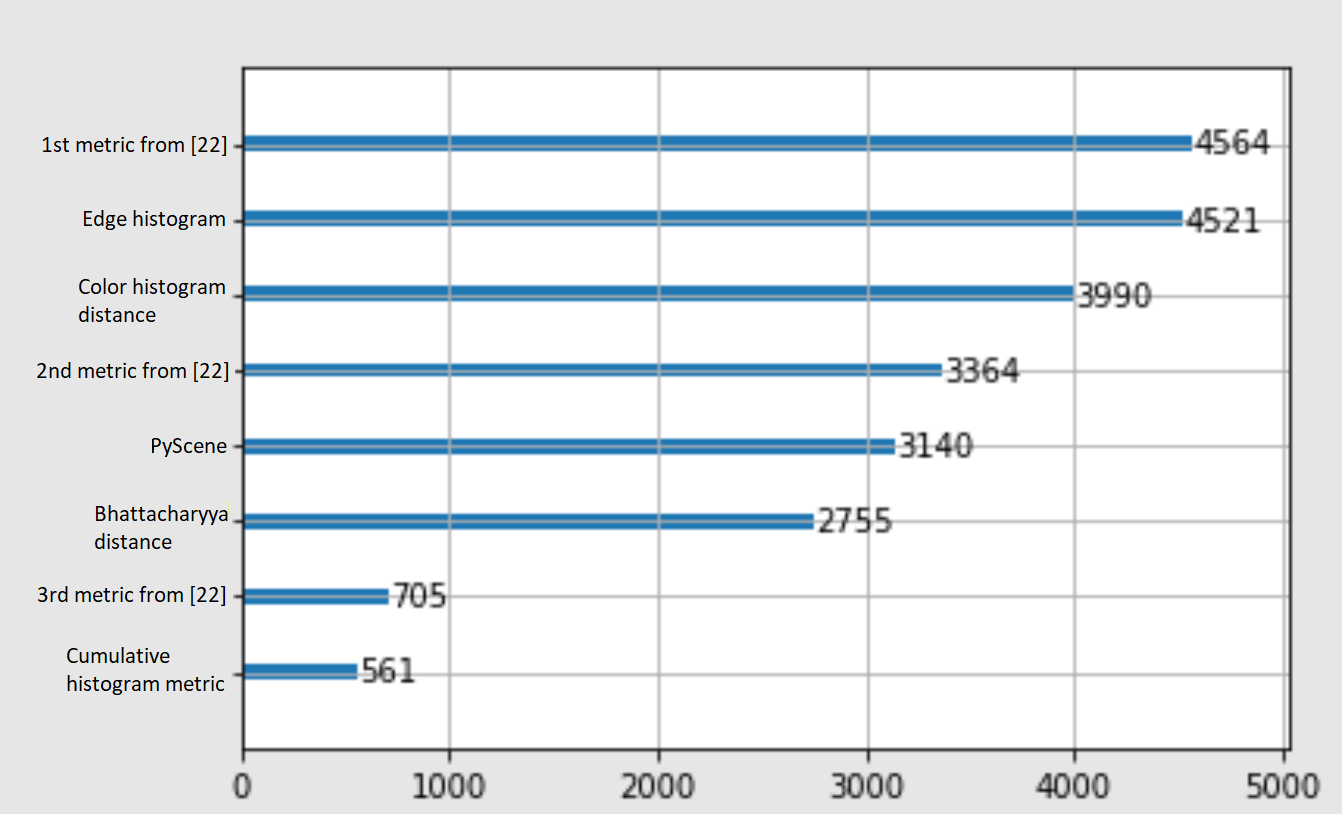}}
\caption{Feature importance.}
\label{fig:1image}
\end{figure}

	As a training dataset were selected 19 videos on video hosting youtube.com total duration of 26 minutes (38917 frames) . Also 2 videos from BBC Planet Earth set were added with total duration of 96 minutes (144700 frames). Thus, the training dataset consisted of 21 videos of 122 minutes duration (183617 frames). There were 917 abrupt scene changes and 54 gradual scene changes.
The test dataset consisted of 9 videos taken from the BBC Planet Earth dataset and 10 videos and the RAI dataset. The total test dataset consisted of 563 minutes of video (804883 frames), with 4510 sharp and 348 gradual scene changes. 
Linear and logistic regression, SVM, K-means, LGBM, and random forest were tried as a learning algorithm. The LGBM algorithm showed the best results, and its parameters were chosen using crossvalidation. The graph of the contribution of the features in the final model can be seen in Fig.~\ref{fig:1image}.

\section{Results}

The accuracy of the algorithm was measured on a test dataset, and a comparison with counterparts was made. The F1 score metric was used to measure the accuracy and recall of the found scene changes. Table~\ref{tab:results} shows the scores obtained: the proposed algorithm outperformed popular methods in terms of accuracy.

\begin{center}
\begin{table}
\caption{Average F1 score for the proposed methods and widely used and popular shot boundary detection methods on BBC Planet Earth and RAI datasets.}
  \label{tab:results}
 \begin{tabular}{ccccc} 
 \toprule
 Method & Speed (FPS) & F score & Precision & Recall \\
 \midrule
 Proposed & 31 & 0.9794 & 0.9784 & 0.9803 \\ 
 \hline
 VQMT\cite{vqmt} & 308 & 0.9476 & 0.9352 & 0.9604 \\
 \hline
 FFmpeg\cite{ff} & 165 & 0.9448 & 0.9307 & 0.9594\\
 \hline
 PyScene\cite{pyscene} & 321 & 0.9526 & 0.9467 & 0.9586 \\ 
 \bottomrule
\end{tabular}
\end{table}
\end{center}

\begin{center}
\begin{table}
\caption{F1 score for the proposed methods and widely used and popular shot boundary detection methods on OS VSD dataset.}
  \label{tab:results}
 \begin{tabular}{cccccc} 
 \toprule
 Method & Speed (FPS) & F score & Precision & Recall \\
 \midrule
 Proposed & 31 & 0.8316 & 0.8212 & 0.8423 \\ 
 \hline
 VQMT\cite{vqmt} & 203 & 0.7379 & 0.7523 & 0.7241 \\
 \hline
 FFmpeg\cite{ff} & 104 & 0.7464 & 0.7571 & 0.7361\\
 \hline
 PyScene\cite{pyscene} & 214 & 0.7560 & 0.7782 & 0.7351 \\ 
 \bottomrule
\end{tabular}
\end{table}
\end{center}

\begin{figure*}[htb]
\center{\includegraphics[width=\linewidth]{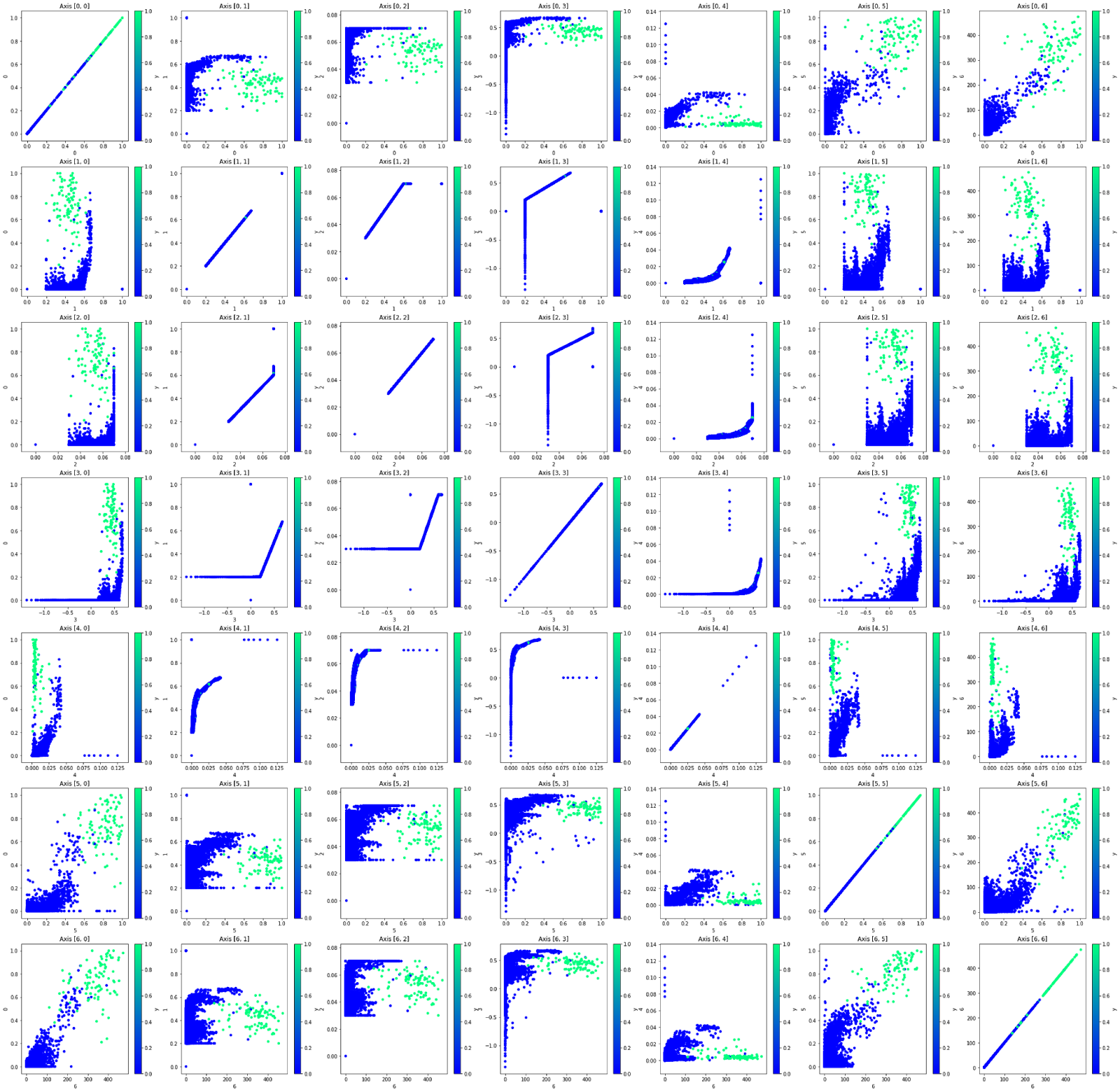}}
\caption{Metrics values on frames from the training dataset. Blue dots--frames without scene changes, green--frames with scene changes}
\label{fig:2image}
\end{figure*}

The efficiency of the used metrics was also analyzed. Fig.~\ref{fig:2image} shows the values of metrics pairs on each frame of training dataset. Different colors indicate the presence and absence of scene changes. From the illustration we can see that many pairs of metrics make a clear classification, for example, a pair of metrics from \cite{dr}.

\begin{figure}[htb]
\center{\includegraphics[width=\linewidth]{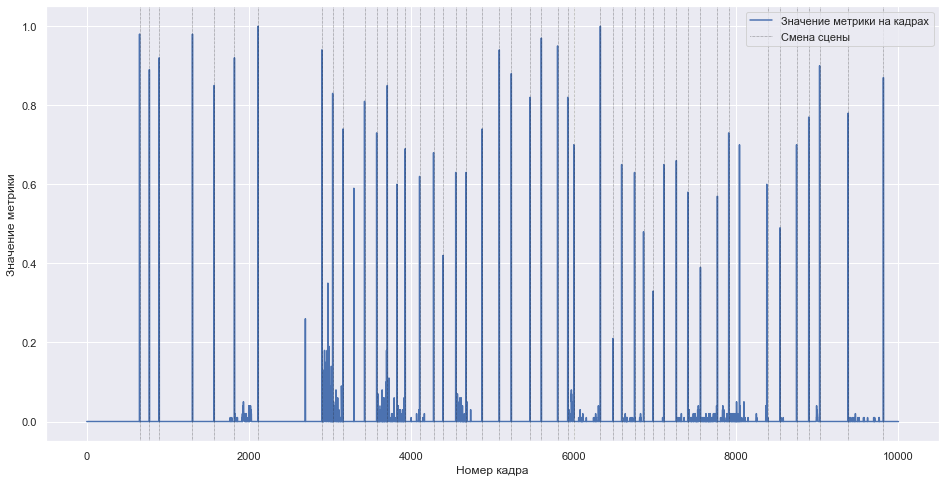}}
\caption{The metric value from the article \cite{dr} on a random segment of the video from the test dataset. Blue line is metric value, vertical lines are frames with scene change.}
\label{fig:image3}
\end{figure}

Fig.~\ref{fig:image3} shows an example of the \cite{dr} metric, which has the largest contribution to the accuracy of the model, on a segment of video from the test set. It can be seen that on the frames with scene changes the metric takes large values, easily separable from the rest.

\section{Conclusion and future work}
In this paper, we proposed a new method for determining scene changes based on different metrics. 
The algorithm was tested on BBC Planet Earth and RAI datasets; its accuracy was 0.9784 and completeness was 0.9803.
The proposed method outperformed its counterparts in the F1-score metric. At the moment the speed of the algorithm is slower than analogues, as it uses a larger number of features. In the further development of the project it is planned to speed up the proposed method, as well as to analyze the performance of the methods in complex cases for classification.

\begin{acknowledgments}
This work was partially supported by the Russian Foundation for Basic Research under Grant 19-01-00785a and the non-commercial fund of science and education development “Intellect”.
\end{acknowledgments}

\balance
\bibliography{output}

\begin{thebibliography}{32}
\expandafter\ifx\csname natexlab\endcsname\relax\def\natexlab#1{#1}\fi
\providecommand{\url}[1]{\texttt{#1}}
\providecommand{\href}[2]{#2}
\providecommand{\path}[1]{#1}
\providecommand{\DOIprefix}{doi:}
\providecommand{\ArXivprefix}{arXiv:}
\providecommand{\URLprefix}{URL: }
\providecommand{\Pubmedprefix}{pmid:}
\providecommand{\doi}[1]{\href{http://dx.doi.org/#1}{\path{#1}}}
\providecommand{\Pubmed}[1]{\href{pmid:#1}{\path{#1}}}
\providecommand{\bibinfo}[2]{#2}
\ifx\xfnm\relax \def\xfnm[#1]{\unskip,\space#1}\fi
\bibitem[{bbc(nd)}]{bbc}
\bibinfo{title}{Bbc planet earth dataset}, \bibinfo{year}{n.d.} \URLprefix
  \url{https://aimagelab.ing.unimore.it/imagelab/research\\Activity.asp?idActivity=19}.
\bibitem[{rai(nd)}]{rai}
\bibinfo{title}{Rai dataset}, \bibinfo{year}{n.d.} \URLprefix
  \url{https://aimagelab.ing.unimore.it/imagelab/research\\Activity.asp?idActivity=19}.
\bibitem[{Chen et~al.(2006)Chen, Shyu, and Zhang}]{first}
\bibinfo{author}{S.-C. Chen}, \bibinfo{author}{M.-L. Shyu},
  \bibinfo{author}{C.~Zhang}, \bibinfo{title}{Innovative Shot Boundary
  Detection for Video Indexing}, \bibinfo{year}{2006}.
  \DOIprefix\doi{10.4018/9781591405719.ch009}.
\bibitem[{H~M et~al.(2020)H~M, Chethan, and B~S}]{second}
\bibinfo{author}{N.~H~M}, \bibinfo{author}{H.~Chethan},
  \bibinfo{author}{R.~B~S},
\newblock \bibinfo{title}{Shot based keyframe extraction using edge-lbp
  approach},
\newblock \bibinfo{journal}{Journal of King Saud University - Computer and
  Information Sciences}  (\bibinfo{year}{2020}).
  \DOIprefix\doi{10.1016/j.jksuci.2020.10.031}.
\bibitem[{thi(nd)}]{third}
\bibinfo{howpublished}{\url{https://ai.googleblog.com/2020/02/autoflip-open-source-framework-for.html}},
  \bibinfo{year}{n.d.}
\bibitem[{Yuan et~al.(2007)Yuan, Wang, Xiao, Zheng, Li, Lin, and
  Zhang}]{formal}
\bibinfo{author}{J.~Yuan}, \bibinfo{author}{H.~Wang},
  \bibinfo{author}{L.~Xiao}, \bibinfo{author}{W.~Zheng},
  \bibinfo{author}{J.~Li}, \bibinfo{author}{F.~Lin},
  \bibinfo{author}{B.~Zhang},
\newblock \bibinfo{title}{A formal study of shot boundary detection},
\newblock \bibinfo{journal}{Circuits and Systems for Video Technology, IEEE
  Transactions on} \bibinfo{volume}{17} (\bibinfo{year}{2007})
  \bibinfo{pages}{168 -- 186}. \DOIprefix\doi{10.1109/TCSVT.2006.888023}.
\bibitem[{H.~Abdulhussain et~al.(2018)H.~Abdulhussain, Ramli, Saripan, Mahmmod,
  Al-Haddad, and Jassim}]{methods}
\bibinfo{author}{S.~H.~Abdulhussain}, \bibinfo{author}{A.~R. Ramli},
  \bibinfo{author}{M.~I. Saripan}, \bibinfo{author}{B.~Mahmmod},
  \bibinfo{author}{S.~A.~R. Al-Haddad}, \bibinfo{author}{W.~Jassim},
\newblock \bibinfo{title}{Methods and challenges in shot boundary detection: A
  review},
\newblock \bibinfo{journal}{Entropy} \bibinfo{volume}{20}
  (\bibinfo{year}{2018}). \DOIprefix\doi{10.3390/E20040214}.
\bibitem[{Boreczky and Rowe(1996)}]{aaa}
\bibinfo{author}{J.~Boreczky}, \bibinfo{author}{L.~Rowe},
\newblock \bibinfo{title}{Comparison of video shot boundary detection
  techniques},
\newblock \bibinfo{journal}{Proceedings of SPIE - The International Society for
  Optical Engineering} \bibinfo{volume}{2670} (\bibinfo{year}{1996}).
  \DOIprefix\doi{10.1117/12.238675}.
\bibitem[{Yoo et~al.(2006)Yoo, Ryoo, and Jang}]{a}
\bibinfo{author}{H.-W. Yoo}, \bibinfo{author}{H.-J. Ryoo},
  \bibinfo{author}{D.-S. Jang},
\newblock \bibinfo{title}{Gradual shot boundary detection using localized edge
  blocks},
\newblock \bibinfo{journal}{Multimedia Tools Appl.} \bibinfo{volume}{28}
  (\bibinfo{year}{2006}) \bibinfo{pages}{283--300}.
  \DOIprefix\doi{10.1007/s11042-006-7715-8}.
\bibitem[{B~S and Nagendraswamy(2021)}]{b}
\bibinfo{author}{R.~B~S}, \bibinfo{author}{H.~Nagendraswamy},
\newblock \bibinfo{title}{Video shot boundary detection using block based
  cumulative approach},
\newblock \bibinfo{journal}{Multimedia Tools and Applications}
  \bibinfo{volume}{80} (\bibinfo{year}{2021}) \bibinfo{pages}{1--24}.
  \DOIprefix\doi{10.1007/s11042-020-09697-6}.
\bibitem[{Sasithradevi and Roomi(2020)}]{c}
\bibinfo{author}{A.~Sasithradevi}, \bibinfo{author}{S.~M.~M. Roomi},
\newblock \bibinfo{title}{A new pyramidal opponent colorshape model based video
  shot boundary detection},
\newblock \bibinfo{journal}{J. Vis. Commun. Image Represent}
  \bibinfo{volume}{67} (\bibinfo{year}{2020}) \bibinfo{pages}{12}.
\bibitem[{Miene et~al.(2002)Miene, Dammeyer, Hermes, and Herzog}]{f}
\bibinfo{author}{A.~Miene}, \bibinfo{author}{A.~Dammeyer},
  \bibinfo{author}{T.~Hermes}, \bibinfo{author}{O.~Herzog},
\newblock \bibinfo{title}{Advanced and adaptive shot boundary detection}
  (\bibinfo{year}{2002}).
\bibitem[{G~G and S(2014)}]{fff}
\bibinfo{author}{L.~P. G~G}, \bibinfo{author}{D.~S},
\newblock \bibinfo{title}{Walsh–hadamard transform kernel-based feature
  vector for shot boundary detection},
\newblock \bibinfo{journal}{IEEE transactions on image processing : a
  publication of the IEEE Signal Processing Society} \bibinfo{volume}{23}
  (\bibinfo{year}{2014}). \DOIprefix\doi{10.1109/TIP.2014.2362652}.
\bibitem[{P.~Panchal(2012)}]{d}
\bibinfo{author}{.~N.~P. P.~Panchal, S.~Merchant},
\newblock \bibinfo{title}{Scene detection and retrieval of video using motion
  vector and occurrence rate of shot boundaries},
\newblock in: \bibinfo{booktitle}{2012 Nirma University International
  Conference on Engineering (NUiCONE)}, volume~\bibinfo{volume}{67},
  \bibinfo{year}{2012}, pp. \bibinfo{pages}{1--6}.
\bibitem[{Jacobs et~al.(2004)Jacobs, Miene, Ioannidis, and Herzog}]{e}
\bibinfo{author}{A.~Jacobs}, \bibinfo{author}{A.~Miene},
  \bibinfo{author}{G.~Ioannidis}, \bibinfo{author}{O.~Herzog},
\newblock \bibinfo{title}{Automatic shot boundary detection combining color,
  edge, and motion features of adjacent frames},
\newblock \bibinfo{year}{2004}, pp. \bibinfo{pages}{197--206}.
\bibitem[{Chakraborty et~al.(2021)Chakraborty, Singh, and Thounaojam}]{bb}
\bibinfo{author}{S.~Chakraborty}, \bibinfo{author}{A.~Singh},
  \bibinfo{author}{D.~Thounaojam},
\newblock \bibinfo{title}{A novel bifold-stage shot boundary detection
  algorithm: invariant to motion and illumination},
\newblock \bibinfo{journal}{The Visual Computer}  (\bibinfo{year}{2021})
  \bibinfo{pages}{1--12}. \DOIprefix\doi{10.1007/s00371-020-02027-9}.
\bibitem[{Amel et~al.(2010)Amel, Ben~Abdelali, and Abdellatif}]{dd}
\bibinfo{author}{A.~Amel}, \bibinfo{author}{A.~Ben~Abdelali},
  \bibinfo{author}{M.~Abdellatif},
\newblock \bibinfo{title}{Video shot boundary detection using motion activity
  descriptor} \bibinfo{volume}{2} (\bibinfo{year}{2010}).
\bibitem[{Baber et~al.(2011)Baber, Afzulpurkar, Dailey, and Bakhtyar}]{g}
\bibinfo{author}{J.~Baber}, \bibinfo{author}{N.~Afzulpurkar},
  \bibinfo{author}{M.~Dailey}, \bibinfo{author}{M.~Bakhtyar},
\newblock \bibinfo{title}{Shot boundary detection from videos using entropy and
  local descriptor}  (\bibinfo{year}{2011}) \bibinfo{pages}{1--6}.
  \DOIprefix\doi{10.1109/ICDSP.2011.6004918}.
\bibitem[{Apostolidis and Mezaris(2014)}]{h}
\bibinfo{author}{E.~Apostolidis}, \bibinfo{author}{V.~Mezaris},
\newblock \bibinfo{title}{Fast shot segmentation combining global and local
  visual descriptors},
\newblock \bibinfo{year}{2014}. \DOIprefix\doi{10.1109/ICASSP.2014.6854873}.
\bibitem[{Tippaya et~al.(2017)Tippaya, Sitjongsataporn, Khan, and
  Chamnongthai}]{j}
\bibinfo{author}{S.~Tippaya}, \bibinfo{author}{S.~Sitjongsataporn},
  \bibinfo{author}{M.~Khan}, \bibinfo{author}{K.~Chamnongthai},
\newblock \bibinfo{title}{Multi-modal visual features-based video shot boundary
  detection},
\newblock \bibinfo{journal}{IEEE Access} \bibinfo{volume}{PP}
  (\bibinfo{year}{2017}) \bibinfo{pages}{1--1}.
  \DOIprefix\doi{10.1109/ACCESS.2017.2717998}.
\bibitem[{Mondal et~al.(2018)Mondal, Kundu, Das, and Chowdhury}]{k}
\bibinfo{author}{J.~Mondal}, \bibinfo{author}{M.~Kundu},
  \bibinfo{author}{S.~Das}, \bibinfo{author}{M.~Chowdhury},
\newblock \bibinfo{title}{Video shot boundary detection using multiscale
  geometric analysis of nsct and least squares support vector machine},
\newblock \bibinfo{journal}{Multimedia Tools and Applications}
  \bibinfo{volume}{77} (\bibinfo{year}{2018}) \bibinfo{pages}{8139--8161}.
  \DOIprefix\doi{10.1007/s11042-017-4707-9}.
\bibitem[{Smeaton et~al.(2010)Smeaton, Over, and Doherty}]{trecvid}
\bibinfo{author}{A.~Smeaton}, \bibinfo{author}{P.~Over},
  \bibinfo{author}{A.~Doherty},
\newblock \bibinfo{title}{Video shot boundary detection: Seven years of trecvid
  activity},
\newblock \bibinfo{journal}{Comput. Vis. Image Underst.} \bibinfo{volume}{114}
  (\bibinfo{year}{2010}) \bibinfo{pages}{411--418}.
\bibitem[{Qian et~al.(2006)Qian, Liu, and Su}]{l}
\bibinfo{author}{X.~Qian}, \bibinfo{author}{G.~Liu}, \bibinfo{author}{R.~Su},
\newblock \bibinfo{title}{Effective fades and flashlight detection based on
  accumulating histogram difference},
\newblock \bibinfo{journal}{IEEE Transactions on Circuits and Systems for Video
  Technology} \bibinfo{volume}{16} (\bibinfo{year}{2006})
  \bibinfo{pages}{1245--1258}. \DOIprefix\doi{10.1109/TCSVT.2006.881858}.
\bibitem[{osv(nd)}]{osvsd}
\bibinfo{title}{Os vsd dataset}, \bibinfo{year}{n.d.} \URLprefix
  \url{https://www.research.ibm.com/haifa/projects/imt/\\video/Video-DataSet.shtml}.
\bibitem[{tol(nd)}]{toloka}
\bibinfo{title}{Yandex.toloka}, \bibinfo{year}{n.d.} \URLprefix
  \url{https://toloka.yandex.ru/}.
\bibitem[{Bogdanov~Alexander(2016)}]{dr}
\bibinfo{author}{B.~J. Bogdanov~Alexander},
\newblock \bibinfo{title}{Adaptive moment detector of instantaneous scene
  changes in a video stream and its training method based on the signs of video
  stream content: dark / light, calm / dynamic},
\newblock in: \bibinfo{booktitle}{Bulletin of Tomsk State University.
  Management, computer science and informatics}, \bibinfo{number}{4 (37)},
  \bibinfo{year}{2016}.
\bibitem[{max(nd)}]{max}
\bibinfo{title}{Maxreimann github repository}, \bibinfo{year}{n.d.} \URLprefix
  \url{https://github.com/MaxReimann/Shot-Boundary-Detection}.
\bibitem[{ays(nd)}]{ays}
\bibinfo{title}{aysebilgegunduz github repository}, \bibinfo{year}{n.d.}
  \URLprefix \url{https://github.com/aysebilgegunduz/ShotBoundary\\Detection}.
\bibitem[{pys(nd)}]{pyscene}
\bibinfo{title}{Pyscene shot boundary detection tool}, \bibinfo{year}{n.d.}
  \URLprefix \url{https://pyscenedetect.readthedocs.io/en/latest/}.
\bibitem[{Murashko et~al.(2016)Murashko, Thomson, and Leather}]{murashko}
\bibinfo{author}{O.~Murashko}, \bibinfo{author}{J.~Thomson},
  \bibinfo{author}{H.~Leather},
\newblock \bibinfo{title}{Predicting and optimizing image compression},
\newblock in: \bibinfo{booktitle}{Proceedings of the 24th ACM International
  Conference on Multimedia}, MM '16, \bibinfo{publisher}{Association for
  Computing Machinery}, \bibinfo{address}{New York, NY, USA},
  \bibinfo{year}{2016}, p. \bibinfo{pages}{665–669}. \URLprefix
  \url{https://doi.org/10.1145/2964284.2967305}.
  \DOIprefix\doi{10.1145/2964284.2967305}.
\bibitem[{vqm(nd)}]{vqmt}
\bibinfo{title}{Msu vqmt scene change detection tool}, \bibinfo{year}{n.d.}
  \URLprefix
  \url{https://www.compression.ru/video/quality-measure/metric-plugins/scd-en.htm}.
\bibitem[{ff(nd)}]{ff}
\bibinfo{title}{Ffmpeg shot boundary detection tool}, \bibinfo{year}{n.d.}
  \URLprefix \url{https://ffmpeg.org/}.

\end{thebibliography}

\end{document}